\documentclass{article}
\usepackage{main}

\usepackage{times}
\usepackage{soul}
\usepackage{url}
\usepackage[hidelinks]{hyperref}
\usepackage[utf8]{inputenc}
\usepackage[small]{caption}
\usepackage{graphicx}
\usepackage{amsmath}
\usepackage{amsthm}
\usepackage{amssymb}
\usepackage{booktabs}
\usepackage{algorithm}
\usepackage{algorithmic}
\usepackage{multirow}   

\newtheorem{theorem}{Theorem}

\title{On Emergences of Non-Classical Statistical Characteristics in Classical Neural Networks}

\author{
Hanyu Zhao\thanks{Equal contribution.} \and
Yang Wu\footnotemark[1] \And
Yuexian Hou\thanks{Corresponding author.}\\
\affiliations
Tianjin University, China\\
\emails
\{hanyzhao, yangwu\_, yxhou\}@tju.edu.cn
}

\begin{document}

\maketitle

\begin{abstract}  
    Inspired by measurement incompatibility and Bell-family inequalities in quantum mechanics, we propose the Non-Classical Network (NCnet), a simple classical neural architecture that stably exhibits non-classical statistical behaviors under typical and interpretable experimental setups. We find non-classicality, measured by the $S$ statistic of CHSH inequality, arises from gradient competitions of hidden-layer neurons shared by multi-tasks. Remarkably, even without physical links supporting explicit communication, one task head can implicitly sense the training task of other task heads via local loss oscillations, leading to non-local correlations in their training outcomes. Specifically, in the low-resource regime, the value of $S$ increases gradually with increasing resources and approaches toward its classical upper-bound 2, which implies that underfitting is alleviated with resources increase. As the model nears the critical scale required for adequate performance, $S$ may temporarily exceed 2. As resources continue to grow, $S$ then asymptotically decays down to and fluctuates around 2. Empirically, when model capacity is insufficient, $S$ is positively correlated with generalization performance, and the regime where $S$ first approaches $2$ often corresponding to good generalization. Overall, our results suggest that non-classical statistics can provide a novel perspective for understanding internal interactions and training dynamics of deep networks.

\end{abstract}

\section{Introduction}
Driven by the concurrent growth of model scale, data availability, and computational resources, Transformer-based architectures(\cite{devlin2019bert,Liu2019RoBERTaAR,achiam2023gpt,dubey2024llama}), particularly large language models(LLMs), have shown strong capabilities in natural language understanding(\cite{wei2025integration,liu2024emollms,zhu2025sentiment}), generation(\cite{mondshine2025beyond,savoldi2025mind,kulkarni-etal-2024-synthdst}), reasoning(\cite{dong-etal-2024-survey,wang2023large}), and cross-task transfer. However, the rapid expansion of model capabilities has also made reliable model evaluation an increasingly prominent challenge. On the one hand, traditional evaluation paradigms rely primarily on single-task benchmarks and performance metrics(\cite{papineni-etal-2002-bleu,van1979information,banerjee2004meteor}), in which score improvements do not necessarily correspond to genuine gains in overall model capability. On the other hand, as application scenarios continue to evolve, evaluation increasingly depends on performance after fine-tuning or instruction alignment. More critically, such metrics typically characterize only the statistical properties of individual tasks and do not provide effective means for assessing internal feature representations or inter-task relationships. Dynamic multi-task performance evaluation is closely aligned with the core requirements of artificial general intelligence (AGI), which posits that a general-purpose agent should be able to transfer flexibly across diverse tasks and their compositions. In practice, such transitions often necessitate partial retraining, typically via parameter fine-tuning.

From the perspective of fine-tuning(\cite{devlin2019bert,houlsby2019parameter,li2021prefix}), when a model is adapted to different tasks, its parameter space often exhibits systematic variations, indicating that each fine-tuned model effectively corresponds to a distinct measurement context within the same semantic space. However, these contexts do not necessarily admit a common global parameter configuration that can satisfy them simultaneously. Different tasks impose optimization objectives(\cite{sener2018multi}) on representations and decision boundaries that may conflict within shared parameter dimensions, such that parameter updates optimized for one task inevitably degrade performance on another. This phenomenon admits a natural analogy to measurement incompatibility(\cite{von2013mathematische,heisenberg1927anschaulichen}) in quantum physics, where non-commuting observables cannot be simultaneously determined and consequently lead to violations of constraints that are otherwise satisfiable under classical assumptions, such as Bell inequalities(\cite{Bell1964OnTE}).

Machine learning techniques have also introduced new analytical paradigms for the study of non-classical correlations. Tamás Kriváchy et al.(\cite{krivachy2020neural}) employed neural networks to reconstruct classical strategies in order to verify nonlocal correlations in quantum networks. Zhang et al.(\cite{zhang2024detecting})developed a detection framework for non-classical correlations in trilocal networks using stacked deep neural networks. However, the effectiveness of existing approaches relies on an implicit assumption that, in the absence of explicit connections, classical feedforward neural networks cannot generate non-classical correlations between outputs, as their operations are describable by local hidden-variable models. We recognize that these assumptions have limitations and propose NCnet, which is designed to stably exhibit non-classical statistical behaviors under specific and interpretable experimental settings. In addition, we introduce the CHSH statistic as an externally observable diagnostic tool, providing a new perspective for understanding internal interaction structures in deep neural networks and for evaluating the performance of large scale models. \textbf{Our contributions are as follows :}
\begin{itemize}
    \item \textbf{Methodological innovation:} We present the first approach that maps the CHSH statistic $S$ onto multi-task models, enabling a quantitative characterization of task cooperation and competition via the perspective of non-classical statistical analysis.
    
    \item \textbf{Architectural Contribution:} We introduce NCnet, a classical neural architecture comprising a shared hidden layer and dual task-specific heads, which consistently exhibits non-classical statistical behavior under well-defined and reproducible experimental conditions.

    \item \textbf{Mechanistic Insight:} We demonstrate that the violation of the CHSH inequality does not arise from any explicit information channel, but is instead driven by gradient competition induced by shared parameters in multi-task learning. Moreover, we show that this phenomenon is most pronounced in a critical regime where model capacity is nearly sufficient yet not redundant.
\end{itemize}

\section{Preliminaries}

\subsection{Local Realism and Local Hidden Variable Models} Locality implies that two spacelike-separated regions are informationally isolated during the considered time interval, as no signal can exceed the speed of light, thereby excluding any form of instantaneous ``action at a distance''. Realism holds that the properties of a physical system, such as position, momentum, or spin, exist objectively prior to measurement and independently of it. Taken together, locality and realism constitute local realism(\cite{einstein1971born,laloe2019we,laudisa2023and}), which stipulates that events in one region cannot immediately affect the physical reality of another, and that any causal influence must respect the finite speed of propagation.

Within the historical debates surrounding the completeness of quantum mechanics, the framework of local hidden variable (LHV) models(\cite{einstein1935can,Bell1964OnTE} ) was proposed under the premise of local realism, representing an attempt to account for quantum phenomena within a classical worldview. Concretely, for a particle $p$ and an observable $O$, the measurement outcome is fully determined by the observable $O$ together with an intrinsic property $\lambda$ of the particle. In other words, when measuring $O$ on particle $p$, the outcome is represented as $O(\lambda)$, where $\lambda$ denotes the local hidden variable associated with $p$.

Within the feedforward neural networks framework considered in this paper, two top-level neurons between which no (explicit) information link exists are regarded as mutually local (i.e., informationally isolated). In this setting, event information occurring at one neuron cannot be transmitted to the other within the time interval under consideration. This definition is logically consistent with the notion of locality as defined in the philosophy of physics.

\subsection{The Classical Upper-Bound of the CHSH Statistic}
To examine whether there exists a fundamental conflict between quantum mechanics and local realism, John Bell proposed the correlation-based Bell inequalities(\cite{Bell1964OnTE}). Clauser, Horne, Shimony, and Holt introduced the CHSH inequality(\cite{clauser1969proposed}), a specific formulation amenable to experimental verification. 

Consider four binary observable variables $A_i,B_j\in \{-1, +1\}$, for $i,j\in \{0, 1\}$, where \( A_0 \) and \( A_1 \) denote the measurement outcomes of one party, Alice, and \( B_0 \) and \( B_1 \) denote the outcomes of the other party, Bob. The experimental setup satisfies the basic conditions of the CHSH inequality: each party independently selects an observable variable in each trial, producing binary outcomes that yield four sets of correlation data.

\begin{theorem}[Classical Upper-Bound of the CHSH Statistic]
\label{thm:chsh_classical}
For any local hidden-variable theory, the CHSH value $S$ satisfies
\begin{equation}
|S| = \left| C_{A_0B_0} + C_{A_0B_1} + C_{A_1B_0} - C_{A_1B_1} \right| \leq 2
\label{eq:CHSHbound}
\end{equation}

where  $C_{A_iB_j} = \mathbb{E}[A_i B_j]$ is an expectation statistic indicates the association between the measurement outcomes of Alice and Bob. If supposing that $A_i$ and $B_j$ are centered and normalized, it turns out that $C_{A_iB_j}$ simply  becomes Person's correlation coefficient.
\end{theorem}

The detailed proof and the corresponding derivation within LHV framework can be found in the literature(\cite{brunner2014bell}). The CHSH inequality provides an operational criterion to distinguish between LHV models and non-classical probabilistic correlations. Theoretically, this inequality must hold for LHV models, as well as for all classical probabilistic models grounded in the assumptions of locality and realism.

\section{Related Work}

\begin{figure*}[t]
    \centering
    \begin{minipage}[b]{0.3\linewidth}
        \centering
        \includegraphics[height=4cm, keepaspectratio]{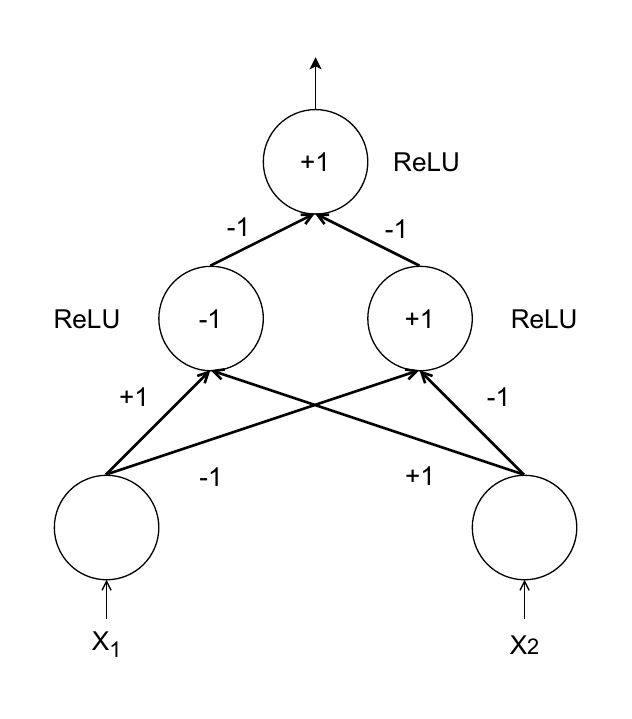}\\
        \small {(a)} XORnet
    \end{minipage}
    \begin{minipage}[b]{0.4\linewidth}
        \centering
        \includegraphics[height=4cm, keepaspectratio]{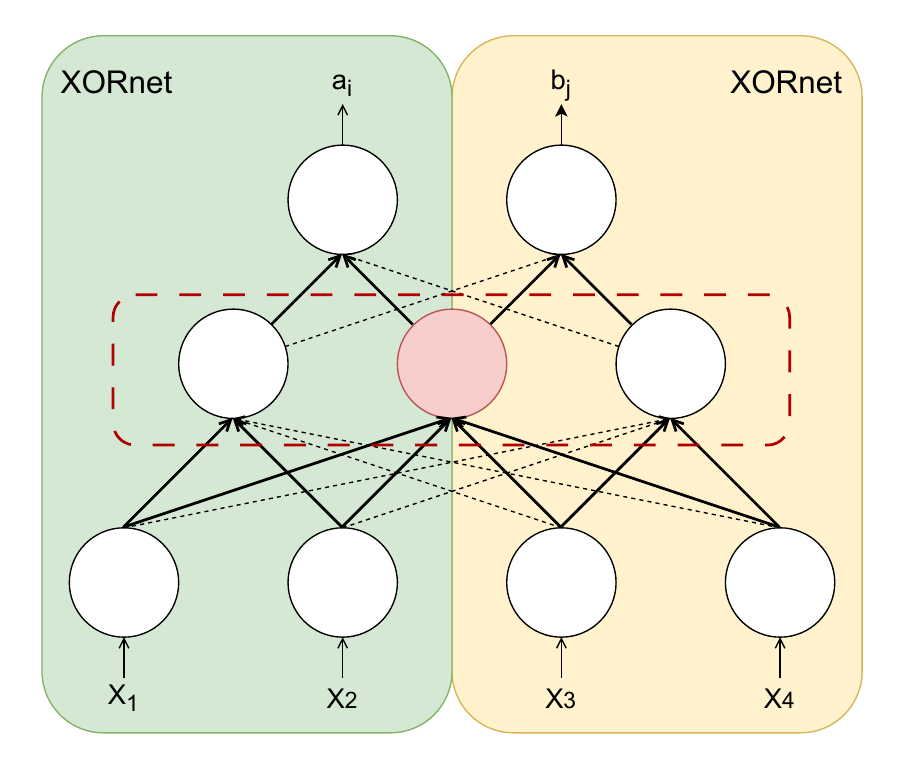}\\
        \small {(b)} NCnet
    \end{minipage}

    \caption{The framework of the proposed NCnet. (a) XORnet illustrates the basic network structure for modeling the XOR function with ReLU activations. (b) NCnet is constructed by integrating two XORnets. The red node denotes a shared neuron where gradient competition is prone to occur.}
    \label{img:xor_ncnet}
\end{figure*}

\textbf{Neural Network–Based Representation and Quantification of Non-Classical Correlations.} Machine learning has emerged as a powerful analytical paradigm for the study of non-classical correlations. Askery Canabarro et al.(\cite{canabarro2019machine}) developed hybrid models combining multilayer perceptrons with genetic algorithms to achieve high-precision quantification of non-classical correlations in Bell scenarios and to discover novel correlation structures.  Luo et al. (\cite{luo2018computationally}) constructed nonlinear Bell inequalities based on bipartite graph matching to characterize multipartite correlations in quantum networks. There also exist some works(\cite{krivachy2020neural,polino2023experimental})  are typically grounded in the assumption that if the correlations present in the data can be perfectly modeled by classical neural networks, then the data contain only classical correlations; conversely, if these correlations cannot be captured by those models, the data are taken to exhibit non-classical correlations. 

\begin{table}[ht]
\begin{center}
\caption{Parameter Mapping Between NCnet and Bell Experiments.}
\label{ncnet:parameter_mapping_table}
\begin{tabular}{p{4cm}l}
\hline
NCnet Parameter & Bell Experiment Parameter \\
\hline
Alice's task $\alpha_i$ & Alice's measurement basis $i$ \\
Bob's task $\beta_j$ & Bob's measurement basis $j$ \\
Consistency indicator $A_i(B_j)$ & Measurement result $A_i(B_j)$ \\
Correlation of measurement outcomes $C(A_i , B_j)$ & Correlation $C_{A_iB_j}$ \\
\hline
\end{tabular}
\end{center}
\end{table}

\section{Methodology}

To more clearly demonstrate the emergence of non-classical statistical phenomena in neural networks, we propose a Non-Classical Network (NCnet) derived from the XORnet. In this paper, the ``non-classical network” is defined as a classical neural network that can robustly exhibit non-classical statistical features. As shown in Figure \ref{img:xor_ncnet}, this architecture provides an insightful and controlled framework for analyzing non-classical phenomena.

\subsection{Framework of NCnet and Task Definition}
\begin{figure*}[t]
    \centering
    \begin{minipage}[b]{0.45\linewidth}
        \centering
        \includegraphics[height=4cm, keepaspectratio]{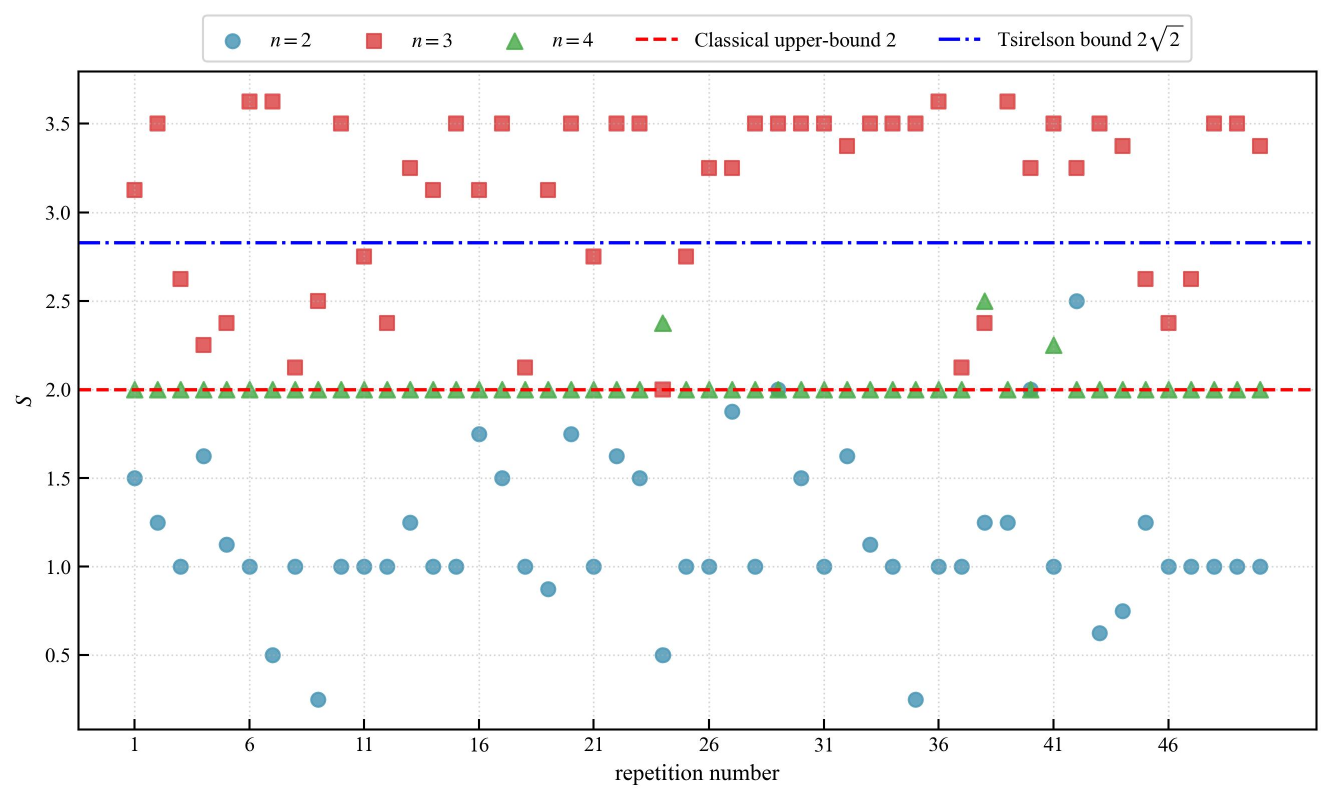}\\
        \small {(a)} Scatter Distribution of $S$
    \end{minipage}
    \begin{minipage}[b]{0.45\linewidth}
        \centering
        \includegraphics[height=4cm, keepaspectratio]{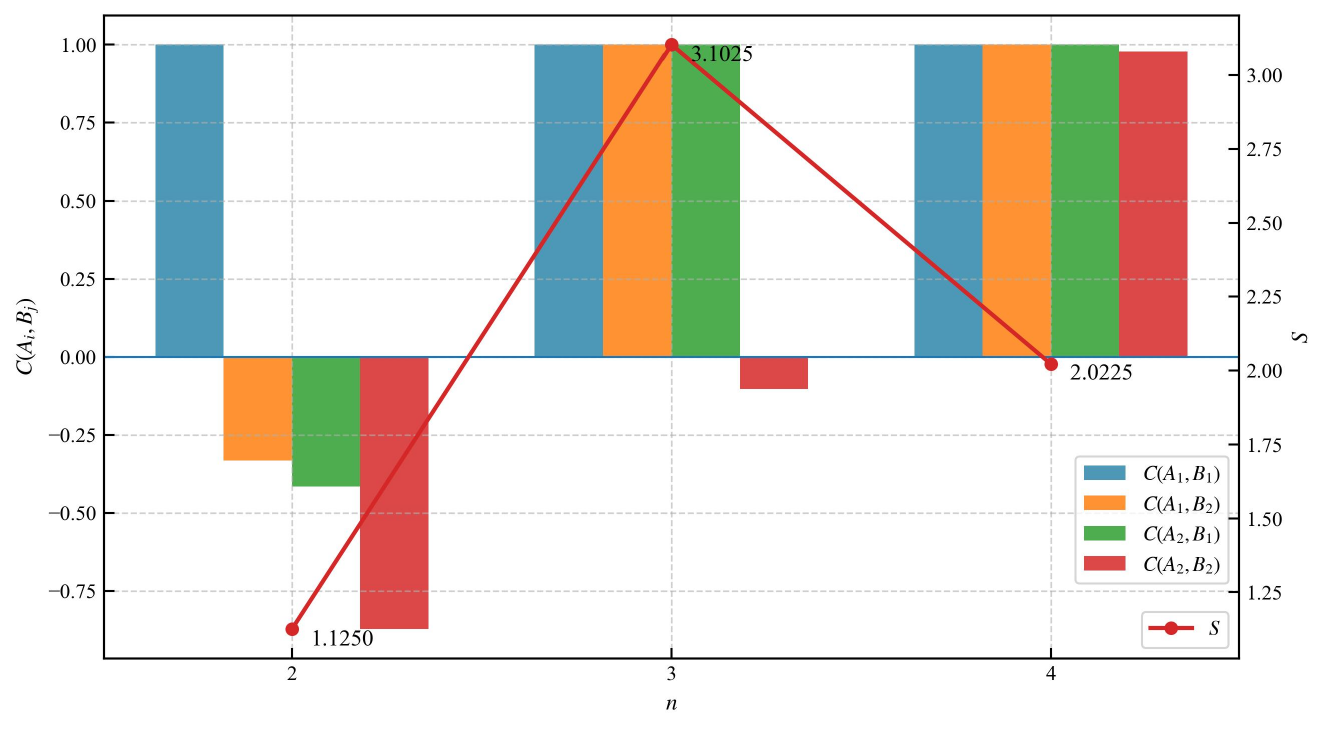}\\
        \small {(b)} Average Correlation Terms and $S$
    \end{minipage}

    \caption{Overall results of NCnet with different hidden-layer sizes. (a) Scatter distribution of $S$ obtained from 50 independent runs of NCnet for hidden-layer sizes $n = 2, 3, 4$. The red dashed line indicates the classical upper-bound of the CHSH statistic, and the blue dot-dashed line marks the Tsirelson bound. (b) Mean correlation values $C(A_i , B_j)$ and the corresponding average statistic $S$ for each value of $n$.}
        
    \label{nc-net:exp}
\end{figure*}

NCnet takes four binary inputs, $X_1$ through $X_4$, after which the network produces two output decisions corresponding to the measurement outcomes at Alice’s and Bob’s sides. We define the tasks on Alice’s side and Bob’s side as follows:
\begin{equation}
\label{eq:task_Alice}
Alice :
\begin{cases}
    \alpha_1 = X_1 \\
   \alpha_2 = X_1 \oplus X_2,
\end{cases}
\end{equation}

\begin{equation}
\label{eq:task_Bob}
Bob :
\begin{cases}
    \beta_1 = X_3 \\
   \beta_2 = X_3 \oplus X_4
\end{cases}
\end{equation}
In this setup, $\alpha_1$ and $\beta_1$ perform identity mappings, while $\alpha_2$ and $\beta_2$ realize XOR logical operations. $A_i$ ($B_j$) denotes the binary measurement outcome as the consistency indicator for task $\alpha_i$ ($\beta_j$), taking value $+1$ if NCnet’s prediction is correct and $-1$ otherwise (similarly for $B_j$). For each task pair $(\alpha_i, \beta_j)$, the correlation of their measurement outcomes is defined as $C(A_i, B_j) = \langle A_i \cdot B_j \rangle, i,j \in \{1,2\}$,
where $\langle \cdot \rangle$ denotes the mathematical expectation, approximated by the arithmetic average in the experiment. The correspondence between NCnet’s prediction tasks and the Bell experimental parameters is summarized in Table~\ref{ncnet:parameter_mapping_table}. Finally, the CHSH statistic $S$ defined as
\begin{equation}
\label{eq:S}
S = C(A_1, B_1) + C(A_1, B_2) + C(A_2, B_1) - C(A_2, B_2)\end{equation}

Indeed, if the random variables $A_i$ and $B_j$ both have a mean of 0 and a standard deviation of 1, then the formula for their correlation $C(A_i,B_j)$ reduces to $\rho_{A_iB_j}$ i.e., Pearson correlation coefficient. If the experiment yields $S > 2$, it indicates the presence of non-classical correlations in the system that cannot be explained by any LHV model.

\subsection{CHSH Inequality Violation Based on NCnet}

To verify the emergence of non-classical features in NCnet under specific conditions, we trained four models corresponding to the four task combinations formulated in Equations~\eqref{eq:task_Alice} and \eqref{eq:task_Bob}, and calculated the statistic $S$. Furthermore, to analyze the impact of model capacity on $S$, we considered three scenarios with the number of hidden neurons set to $n=2,3,4$. All experiments were repeated 50 times to ensure stability and reliability.

As shown in Figure~\ref{nc-net:exp}, the statistic $S$ exhibits a pronounced nonlinear dependence on the number of hidden units $n$. When $n = 2$, the values of $S$ are generally low: in most runs $S < 1.5$, which is well below the LHV bound of $2.0$. When $n = 3$, $S$ attains its maximum. Almost all repeated runs yield values that significantly exceed the LHV upper-bound. In some cases, the observed values even surpass the Tsirelson bound(\cite{cirel1980quantum}) $2\sqrt{2} \approx 2.828$, reaching approximately $S \approx 3.5$. When $n = 4$, $S$ decreases and stabilizes around $2$, and the apparent violation of CHSH inequality disappears. In summary, the nonlinear relationship between $S$ and the number of hidden layer neurons $n$ may represent a universally applicable phenomenon. Specifically, the most pronounced non-classical features emerge near a critical regime in which the network’s representational capacity is nearing an adequate level, while remaining insufficient.

\subsection{Causes of Non-Classical Features}
In the experiments described above, when the number of neurons was set to three, the statistic $S$ exceeded the LHV upper-bound in all 50 repeated trials. Formally, this behavior is reminiscent of ``nonlocal'' correlations. When Alice and Bob perform any task combination other than $\alpha_2$ and $\beta_2$, the hidden layer of NCnet is capable of representing the features required by all necessary tasks; that is, there exists at least one set of weights that allows NCnet to achieve perfect training performance on these combinations. In contrast, when Alice and Bob perform the task combination $\alpha_2$ and $\beta_2$, the hidden layer of NCnet fails to capture at least one of the necessary features, meaning that no set of weights can be found that enables NCnet to train perfectly on this combination. Under this condition, the first three terms of the CHSH statistic $S$ are close to $1$,  whereas the last term is far smaller than $1$, which leads to the emergence of non-classical statistical features. 

Here the fundamental reason for non-classicality is that the top-layer neurons realize implicit communication via gradient competition in the absence of an explicit information-transmission pathway. Gradient competition occurs during backpropagation when a hidden-layer neuron receives conflicting gradient updates from two top-layer neurons. Note that gradient competition is inevitable if the number of neurons in the feature representation layer is insufficient. Persistent gradient competition induces oscillations in the loss functions associated with the top-layer neurons, impeding convergence.  In principle, an observer confined to a single top-layer neuron could infer that the other top-layer neuron is attempting to optimize a more difficult training task by monitoring oscillations in its local training cost function. This constitutes the mechanism by which implicit communication emerges.

\begin{table*}[!htbp]
\centering
\setlength{\tabcolsep}{10pt} %
\caption{Accuracy Acc$_{A_i}$(Acc$_{B_j}$), Correlation $C(A_i, B_j)$, and CHSH Statistics $S$ for Multilingual Training and Mixed Reasoning Tasks across Varying LoRA Ranks $r$.}
\label{tab:exp1_exp2_result}
\begin{tabular}{c|c|ccc|c||ccc|c}
\hline
\multirow{2}{*}{\textbf{$r$}} &
\multirow{2}{*}{\textbf{Task}} &
\multicolumn{4}{c||}{\textbf{Multilingual Training}} &
\multicolumn{4}{c}{\textbf{Mixed Reasoning Tasks}} \\
\cline{3-10}
& & Acc$_{A_i}$ & Acc$_{B_j}$ & $C(A_i, B_j)$ & \textbf{$S$}
  & Acc$_{A_i}$ & Acc$_{B_j}$ & $C(A_i, B_j)$ & \textbf{$S$} \\
\hline

\multirow{4}{*}{1} & $A_1B_1$ & 95.44\% & 91.31\% & 0.7501  & \multirow{4}{*}{1.502}
                              & 99.68\% & 99.82\% & 0.9899 & \multirow{4}{*}{1.902} \\
& $A_1B_2$ & 95.65\% & 89.34\% & 0.7192 &  & 98.92\% & 76.16\% & 0.5144 & \\
& $A_2B_1$ & 93.68\% & 91.28\% & 0.7202 &  & 81.79\% & 98.88\% & 0.6197 & \\
& $A_2B_2$ & 93.80\% & 89.18\% & 0.6871 &  & 79.89\% & 68.91\% & 0.2219 & \\
\hline

\multirow{4}{*}{2} & $A_1B_1$ & 97.32\% & 94.67\% & 0.8453 & \multirow{4}{*}{1.713}
                              & 99.92\% & 99.96\% & 0.9976 & \multirow{4}{*}{2.191} \\
& $A_1B_2$ & 97.26\% & 94.41\% & 0.8445 &  & 99.46\% & 93.89\% & 0.8670 & \\
& $A_2B_1$ & 96.31\% & 95.35\% & 0.8397 &  & 96.71\% & 99.66\% & 0.9299 & \\
& $A_2B_2$ & 96.29\% & 93.91\% & 0.8161  &  & 93.33\% & 84.58\% & 0.6038 & \\
\hline

\multirow{4}{*}{4} & $A_1B_1$ & 98.79\% & 97.83\% & 0.9335  & \multirow{4}{*}{1.870}
                              & 99.98\% & 99.99\% & 0.9996 & \multirow{4}{*}{2.103} \\
& $A_1B_2$ & 98.75\% & 97.21\% & 0.9205 &  & 99.94\% & 97.38\% & 0.9463 & \\
& $A_2B_1$ & 97.97\% & 97.64\% & 0.9141 &  & 99.58\% & 99.96\% & 0.9908 & \\
& $A_2B_2$ & 97.96\% & 96.82\% & 0.8983  &  & 97.48\% & 93.77\% & 0.8337 & \\
\hline

\multirow{4}{*}{8} & $A_1B_1$ & 99.50\% & 98.88\% & 0.9678  & \multirow{4}{*}{1.931}
                              & 99.98\% & 99.99\% & 0.9995 & \multirow{4}{*}{2.001} \\
& $A_1B_2$ & 99.32\% & 98.40\% & 0.9547 &  & 99.90\% & 99.54\% & 0.9887 & \\
& $A_2B_1$ & 98.98\% & 98.79\% & 0.9557 &  & 99.88\% & 99.98\% & 0.9971 & \\
& $A_2B_2$ & 98.87\% & 98.50\% & 0.9483 &  & 99.86\% & 99.34\% & 0.9839 & \\
\hline

\multirow{4}{*}{16} & $A_1B_1$ & 99.51\% & 98.90\% & 0.9692  & \multirow{4}{*}{1.938}
                               & 99.72\% & 99.62\% & 0.9867 & \multirow{4}{*}{2.020} \\
& $A_1B_2$ & 99.44\% & 98.73\% & 0.9633 &  & 99.96\% & 99.86\% & 0.9963 & \\
& $A_2B_1$ & 98.97\% & 98.84\% & 0.9571 &  & 99.98\% & 99.98\% & 0.9995 & \\
& $A_2B_2$ & 98.94\% & 98.60\% & 0.9515  &  & 99.44\% & 98.68\% & 0.9623 & \\
\hline

\multirow{4}{*}{32} & $A_1B_1$ & 99.94\% & 99.68\% & 0.9931  & \multirow{4}{*}{1.985}
                              & 99.66\% & 99.56\% & 0.9844 & \multirow{4}{*}{2.021} \\
& $A_1B_2$ & 99.96\% & 99.66\% & 0.9903 &  & 99.96\% & 99.76\% & 0.9944 & \\
& $A_2B_1$ & 99.72\% & 99.60\% & 0.9863 &  & 98.72\% & 99.64\% & 0.9671 & \\
& $A_2B_2$ & 99.64\% & 99.70\% & 0.9867  &  & 98.78\% & 97.38\% & 0.9247 & \\
\hline

\end{tabular}
\end{table*}

This finding carries significant research implications. Given that feedforward neural networks constitute fundamental building blocks of modern deep learning systems and are widely employed across diverse tasks and architectures, it is reasonable to hypothesize that \textbf{non-classical statistical features may be pervasive in contemporary deep learning models.} Moreover, \textbf{Bell inequalities and their generalized forms (i.e., the family of Bell inequalities) provide a theoretical foundation for establishing a novel analytical framework for neural networks.}
This framework not only reveals implicit coupling relationships among different tasks or modules but also serves as a valuable complement to conventional methods of performance evaluation, offering a new analytical perspective for understanding the internal representational mechanisms and training dynamics of neural networks.

\section{Real-World Experiments}

This experiment aims to further verify that non-classical features may also emerge within complex neural networks, and to explore the introduction of the CHSH statistic as a novel model evaluation metric. This metric serves to quantitatively assess a neural network’s representational capacity and generalization performance. Unlike the earlier model based on NCnet, this section focuses on multi-layer architectures that more closely resemble real-world task scenarios, such as contemporary large language models. The investigation examines how non-classical correlations manifest in higher-dimensional and more complex learning tasks.

\subsection{Experiments Setup}
\textbf{Model Architecture and Datasets.} We adopt Multilingual BERT (mBERT) (\cite{pires2019multilingual}) and BERT as the base models, both consisting of 12 Transformer encoder layers, 12 self-attention heads, and 768-dimensional hidden representations. 
Following the previously described NCnet architecture, we adopt a standard multi-task learning framework with hard parameter sharing, in which independent task-specific heads are attached to a shared semantic representation layer, corresponding to the observational outputs of Alice and Bob. To enable fine-grained control over model capacity and support efficient multi-task adaptation, we inject trainable low-rank matrices(\cite{hu2022lora}) into the query and value modules of the self-attention mechanism, adjusting the rank $r \in \{1, 2, 4, 8, 16, 32, 64, 128, 256, 512\}$ to modulate the effective capacity. This allows us to systematically examine how model size influences coupling and competition among tasks during training. Throughout this paper, model capacity refers to the total number of trainable parameters introduced by the LoRA modules together with those in the fully connected layers of the task-specific heads. The specific number of trainable parameters for $r$ ranging from 1 to 32 is provided in Table~\ref{tab:parameter_statistics}. In the following experiments, we construct a multi-task setting using the PAWS-X(\cite{yang2019paws}), SST-2(\cite{wang-etal-2018-glue}, MRPC, CommonsenseQA(\cite{Talmor2019CommonsenseQAAQ}), and MathQA(\cite{Amini2019MathQATI}) datasets. The specific task assignments for Alice and Bob are summarized in the following Table \ref{tab:task_combinations}.

\begin{table}[!htbp]
\centering
\caption{Different Task Combinations of Experiments}
\label{tab:task_combinations}
\begin{tabular}{c c c}  %
\hline
\textbf{Task} & \textbf{Multilingual Training} & \textbf{Mixed Reasoning Tasks} \\
\hline
$A_1$ & PAWS-X-en & SST-2 \\
$A_2$ & PAWS-X-fr & CommonsenseQA \\
$B_1$ & PAWS-X-ja & MRPC \\
$B_2$ & PAWS-X-ko & MathQA \\
\hline
\end{tabular}
\end{table}

\noindent\textbf{Statistical Definitions.}
Within this framework, we adopt the previously defined correlation measure $C(A_i,B_j)$ and the CHSH statistic $S$, applying them to the subsequent experimental analysis. The measure  $C(A_i,B_j)$ quantifies task compatibility: values approaching +1 indicate strong positive correlation between tasks. Conversely, when the value significantly falls below 1, it implies competition or interference between tasks. The absolute value of the CHSH statistic $S$ can serve as a sufficient criterion for the presence of non-classicality. When $\lvert S \rvert$ significantly exceeds 2, it provides evidence for the robust existence of non-classical correlations.

\begin{figure}[t]
\centering
\includegraphics[width=0.35\textwidth]{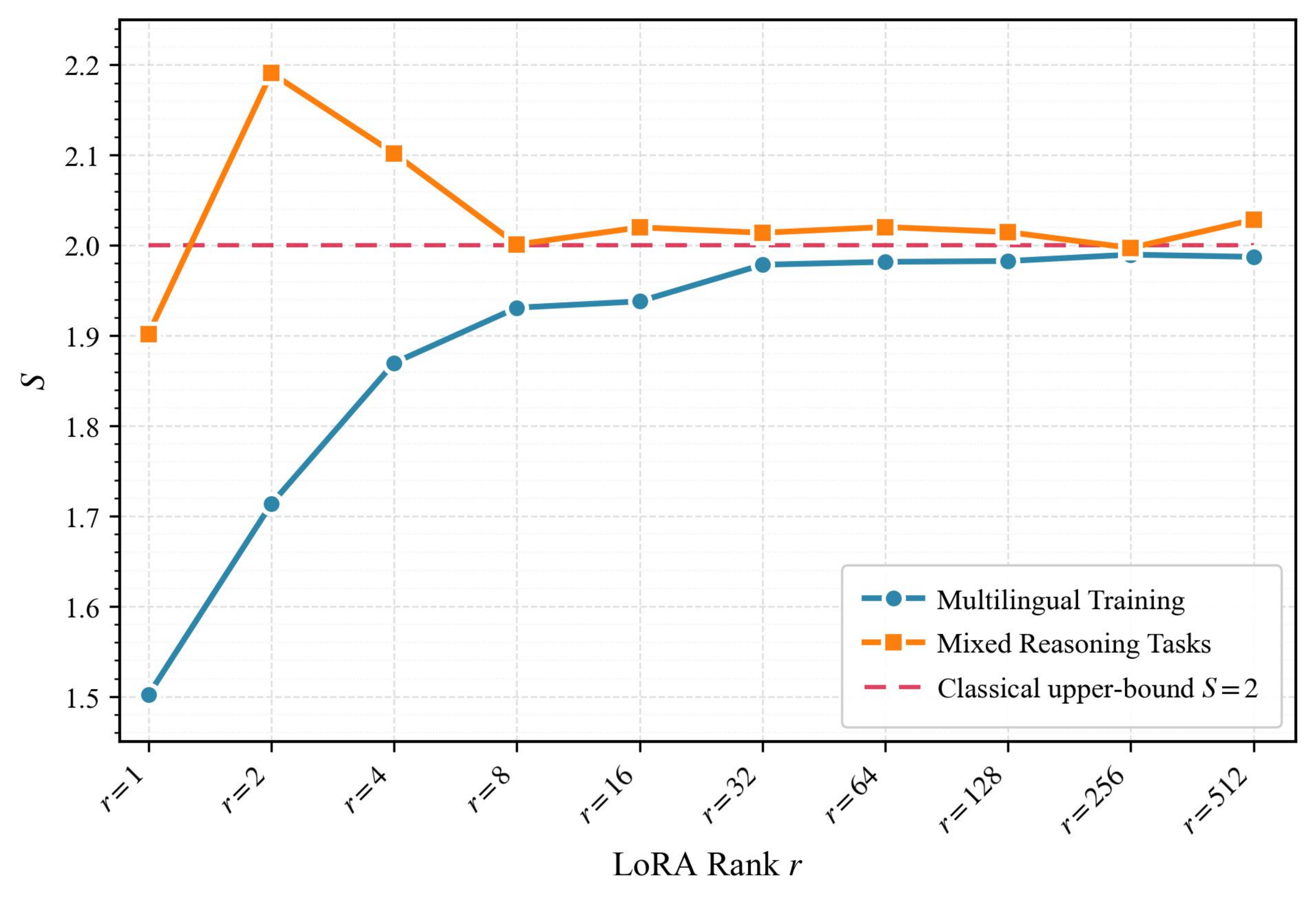}
\caption{The trend of the CHSH statistic $S$ as a function of the rank $r$. The plot includes two sets of experimental results: the blue curve (Multilingual Training) and the orange curve (Mixed Reasoning Tasks). The red dashed line represents the classical upper-bound $S = 2$.
}
\label{S-r}
\end{figure}

\begin{figure*}[htbp]
    \centering
    \begin{minipage}[b]{0.45\textwidth}
        \centering
        \includegraphics[height=4cm, keepaspectratio]{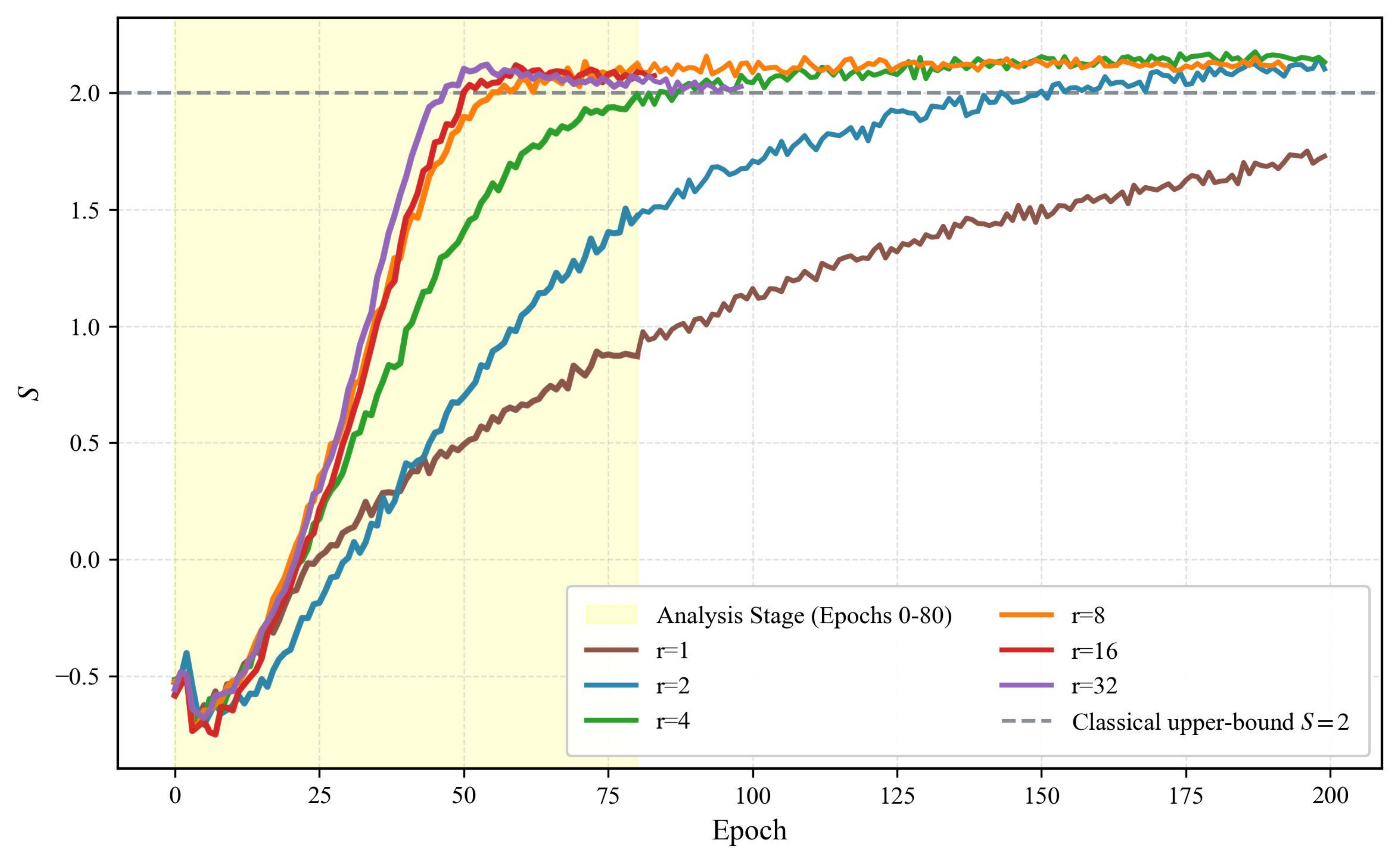}\\
         \small {(a) Evolution of $S$ over Training Epochs} 
    \end{minipage}
    \begin{minipage}[b]{0.45\textwidth}
        \centering
         \includegraphics[height=4cm, keepaspectratio]{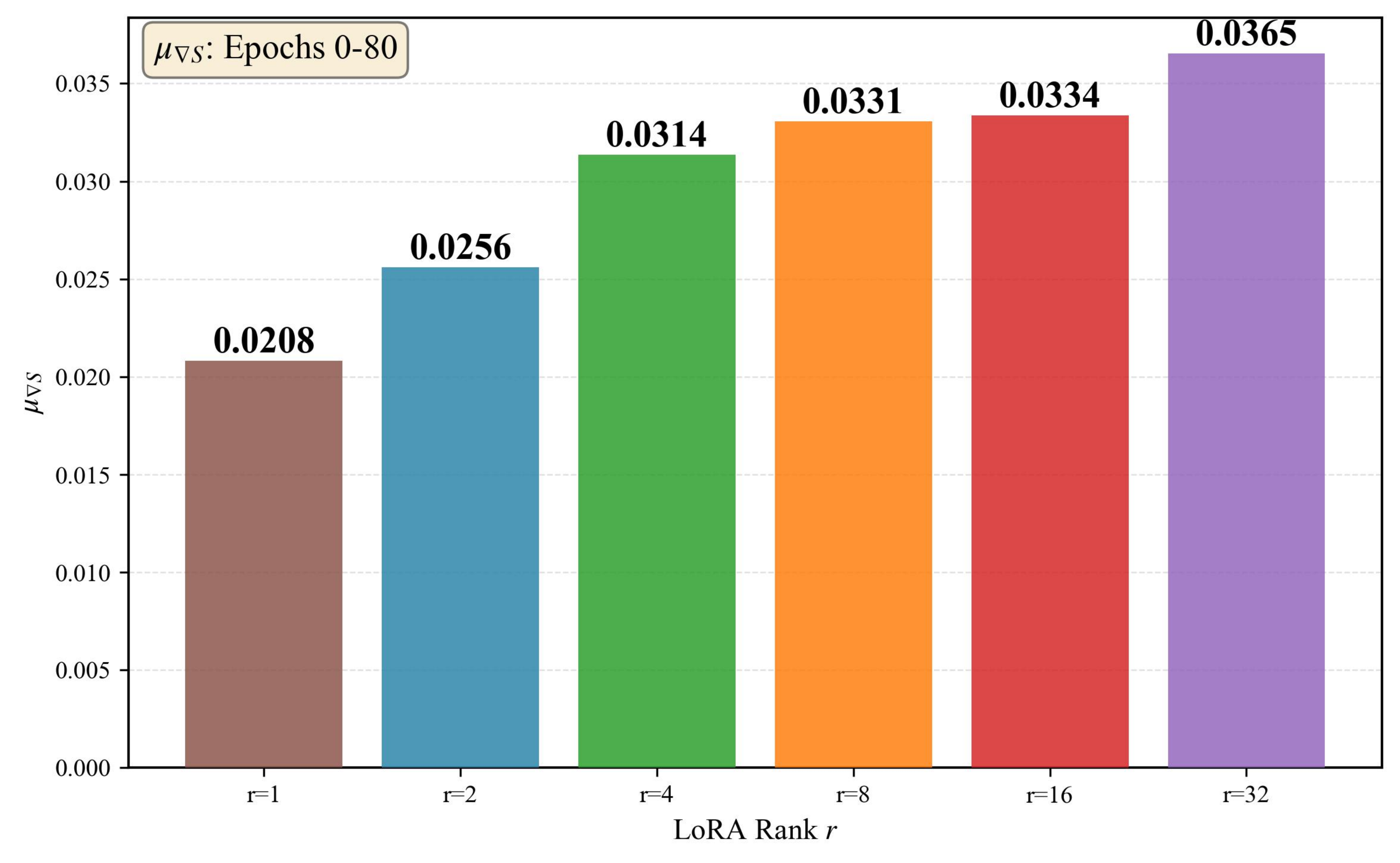}\\
         \small {(b) Convergence Rate}
    \end{minipage}
    \hfill
    \caption{The CHSH statistic $S$ convergence across different ranks under Mixed Reasoning Tasks. (a) Higher ranks $r$ lead to faster convergence of $S$. (b) The bar chart reports $\mu_{\nabla S}$, the arithmetic mean of the instantaneous slopes of $S$ over epochs 0–80; larger bars indicate a faster convergence rate.}
    \label{exp2}
\end{figure*}

\subsection{Experiments Results}
We trained models based on the settings in Table~\ref{tab:task_combinations} to compute the measurement correlations $C(A_i, B_j)$ and the CHSH statistic $S$. Detailed numerical results are provided in Table~\ref{tab:exp1_exp2_result}. Figure~\ref{S-r} illustrates the variation of the CHSH statistic $S$ with the LoRA rank $r$. In the Mixed Reasoning Tasks, we observe a significant violation of the classical upper-bound ($S=2$) at low ranks ($r=2$ and $r=4$), where $S$ reaches its peak. As $r$ increases beyond 4, $S$ declines and asymptotically converges to 2. In contrast, within the Multilingual Training scenario, the value of $S$ increases monotonically as $r$ increases and gradually converges to 2, with no instances significantly exceeding 2 observed. In this setting, the four task combinations exhibit relatively balanced difficulty levels, which suggests that non-classical correlations induced by gradient competition during multi-task training may be moderated. This hypothesis is clearly supported by the results in the low-rank region of Table~\ref{tab:exp1_exp2_result}, where the $C(A_i, B_j)$ values across different task combinations remain numerically close. This explains why $S$ does not significantly exceed 2.

The experimental results on Mixed Reasoning Tasks demonstrate that, because the difficulty of different task combinations varies, classical neural networks for real-world tasks can exhibit pronounced non-classical statistical characteristics, significantly violating the CHSH inequality. This observation suggests that the non-classicality captured by the CHCH statistic $S$ may be pervasive since real-world tasks typically exhibit varying levels of difficulty. Note also that, because $S>2$ represents only a sufficient, rather than a necessary, condition for the existence of non-classicality, there may in principle exist alternative criteria for identifying non-classical behavior that correspond to different non-classicality generation scenarios. Consequently, this observation offers a new perspective that motivates a re-examination of the implicit assumptions underlying certain models for distinguishing non-classical correlations, particularly the assumption that classical neural networks are incapable of generating non-classical correlations.

\subsection{Experimental Summary and Discussion}
\begin{figure}[t]
\centering
\includegraphics[width=\linewidth]{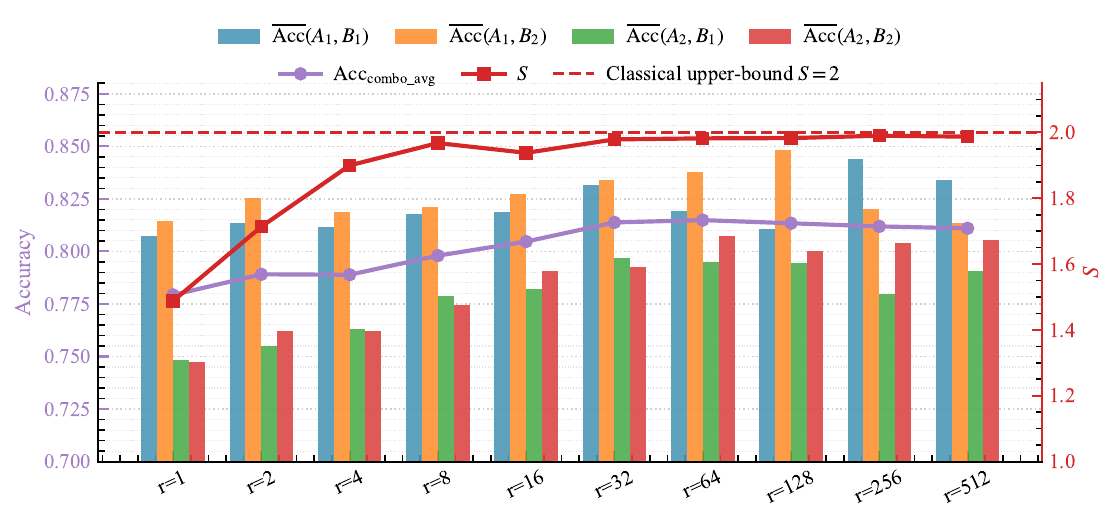}
\caption{
Generalization performance and the CHSH statistic $S$ across ranks $r$ in Multilingual Training. Bars show task-pair mean accuracy $\overline{\mathrm{Acc}}(A_i,B_j)=\frac{\mathrm{Acc}(A_i)+\mathrm{Acc}(B_j)}{2}$. The purple curve shows the combination average $\mathrm{Acc}_{\mathrm{comb\_avg}}$ (the mean of the four bars) per rank. The red curve denotes $S$ value, reflecting the non-classical coupling strength of their learned representations.}
\label{S-gene}
\end{figure}
\subsubsection{Dynamic Analysis of the Training Process}
 In this section, we investigate the relationship between the LoRA rank parameter and the average gradient of the CHSH statistic $S$ during multi-task training. The average gradient is computed as the arithmetic mean of the instantaneous slopes of $S$ across training epochs. As shown in Figure~\ref{exp2}, when $r=1$, the average gradient of $S$ attains its minimum value, with $\mu_{\nabla S} = 0.0208$. As $r$ increases from 2 onward, the average gradient exhibits an overall upward trend. In the range $r=8$ to $r=32$, although the rate of increase diminishes, the average gradient remains at a relatively high level. Overall, the average gradient of the CHSH statistic $S$ is positively correlated with the trainable parameter capacity introduced by LoRA. As $r$ increases, $S$ converges more rapidly during training. A plausible explanation is that higher ranks introduce additional trainable parameters, thereby enhancing the model’s representational capacity and alleviating gradient competition, which in turn improves the efficiency of cross-task gradient propagation.

\subsubsection{Generalization Ability and Non-classical Correlation Analysis under Different LoRA Ranks} 
In this section, we investigate the relationship between the model's generalization capability and the CHSH statistic $S$. The generalization capability is characterized by the model's testing accuracy. From the Figure~\ref{S-gene} , we observe a strong positive correlation between the model's generalization capability and the statistical indicator $S$ within the range $1\leq r \leq 32$. The model capacity at which $r$ first approaches 2 corresponds to a scale that is sufficient but not redundant. According to classical statistical learning theory (e.g., the bias-variance trade-off), this specific model capacity is expected to yield near-optimal generalization performance.

\begin{table*}[!htbp]
\centering
\setlength{\tabcolsep}{12pt}
\caption{Parameter Statistics for Different LoRA Rank Parameters $ r $ Across Experiments}
\label{tab:parameter_statistics}
\begin{tabular}{c|rrr|rrr}
\hline
\multirow{2}{*}{\textbf{$r$}} & 
\multicolumn{3}{c|}{\textbf{Multilingual Training}} & 
\multicolumn{3}{c}{\textbf{Mixed Reasoning Tasks}} \\
\cline{2-7}
& \textbf{\begin{tabular}[c]{@{}c@{}}Total\\Params\end{tabular}} 
& \textbf{\begin{tabular}[c]{@{}c@{}}Trainable\\Params\end{tabular}} 
& \textbf{\begin{tabular}[c]{@{}c@{}}Trainable\\Ratio (\%)\end{tabular}} 
& \textbf{\begin{tabular}[c]{@{}c@{}}Total\\Params\end{tabular}} 
& \textbf{\begin{tabular}[c]{@{}c@{}}Trainable\\Params\end{tabular}} 
& \textbf{\begin{tabular}[c]{@{}c@{}}Trainable\\Ratio (\%)\end{tabular}} \\
\hline
1   & 177,893,380 & 39,940    & 0.022 & 109,522,180 & 39,940    & 0.036 \\
2   & 177,930,244 & 76,804    & 0.043 & 109,559,044 & 76,804    & 0.070 \\
4   & 178,003,972 & 150,532   & 0.085 & 109,632,772 & 150,532   & 0.137 \\
8   & 178,151,428 & 297,988   & 0.167 & 109,780,228 & 297,988   & 0.271 \\
16  & 178,446,340 & 592,900   & 0.332 & 110,075,140 & 592,900   & 0.539 \\
32  & 179,036,932 & 1183,492  & 0.661 & 110,664,964 & 1,182,724 & 1.070 \\
\hline
\end{tabular}
\end{table*}

However, in the Mixed Reasoning task, the model exhibits consistently low average testing accuracy across sub-tasks, with no significant variation as $r$ increases. This phenomenon is attributed to the high complexity and open-ended nature of the tasks. Consequently, testing performance is statistically independent of the model capacity, implying that there is also no statistical correlation between $S$ and generalization capability at the scale of the model under consideration.

\subsubsection{Resource Competition Hypotheses for Non-classicality Witnessed by the CHSH Statistic S} To further elucidate the mechanism by which variations in the LoRA rank parameter influence the CHSH statistic $S$, we propose the following hypotheses based on our experimental observations.

\textbf{Assumption 1:} Different task pairs $(A_i, B_j)$ exhibit heterogeneous demands for model resources (e.g., trainable parameters). Specifically, certain task pairs achieve effective coordination with limited resources, whereas others require substantially higher parameter capacity to converge.

\textbf{Assumption 2:} Under resource-limited conditions, subtasks within a task pair compete for shared resources. This competition increases the possibility that the product $A_i \cdot B_j$ takes negative values, thereby reducing the magnitude of $C(A_i,B_j)$.

\textbf{Assumption 3:} When resources become abundant, inter-task competition is substantially mitigated. Consequently, $A_i$ and $B_j$ tend to take positive values simultaneously, driving $C(A_i,B_j)$ close to $1$.

Based on the above hypothesis and experimental results, we draw the following conclusion: 
\begin{itemize}
  \item \textbf{When $S \ll 2$\footnote{In this paper, the symbols ``$\ll$'' and ``$\gg$'' are used in a qualitative sense to indicate that $S$ is significantly smaller or larger than the bound $2$.}, the model is underfitting and performs poorly across nearly all task combinations, indicating insufficient model capacity or limited ability to form shared representations.}

  \item \textbf{When $S \approx 2$ and $C(A_i,B_j)$ for all task pairs are close to 1, the model has converged well across all combinations.} This suggests that the trainable parameter capacity is well-matched to the representational demands, potentially even exhibiting redundancy.

  \item \textbf{When $S \gg 2$, the model has well converged across most task pairs but fails to fully converge on one specific combination due to persistent gradient competition.} In this regime, salient non-classical statistical features emerge, suggesting that the model operates near a critical point where representational capacity is approaching sufficient, but remains insufficient.
\end{itemize}

\section{Conclusion}
In conclusion, this paper introduces NCnet, a classical neural network architecture demonstrated to robustly exhibit non-classical statistical characteristics within the theoretical framework of measurement (probabilistic) incompatibility. The emergence of these non-classical correlations arises from implicit communication between top-layer neurons in the absence of explicit information exchange pathways. An observer confined to a single top-layer neuron could infer, solely from oscillations in the neuron’s local training loss, that the other top-layer neuron is engaged in a more challenging optimization task.

Through extended experiments on complex architectures and real-world datasets, we further validate the generality and stability of this phenomenon. In parallel, we propose employing the CHSH inequality as a useful analytical metric for neural networks. Note that the CHSH statistic $S$ is only one of the numerous statistical measures for deciding the existence of non-classical statistical features. In principle, we can select or construct appropriate deciding statistics based on the specific problem settings of the task at hand, which indicates a general applicability of non-classical statistical methods in the performance analysis of deep learning models. Unlike conventional metrics based on performance or loss, this approach reveals the internal information-coupling structure of neural systems by quantitatively analyzing the statistical behavior of observable outputs.

\bibliographystyle{named}
\bibliography{main}

\end{document}